\documentclass[a4paper,fleqn]{cas-sc}

\usepackage{booktabs}
\usepackage{graphicx}
\usepackage{color}
\usepackage{algorithm}
\usepackage{algorithmic}
\usepackage{bm}
\usepackage{url}
\usepackage{amssymb}
\usepackage{amsmath}

\begin{document}
\let\WriteBookmarks\relax
\def\floatpagepagefraction{1}
\def\textpagefraction{.001}
\shorttitle{Domain-invariant Clinical Representation Learning by Bridging Data Distribution Shift across EMR Datasets}
\shortauthors{Zhongji Zhang et~al.}

\title [mode = title]{Domain-invariant Clinical Representation Learning by Bridging Data Distribution Shift across EMR Datasets}  

\author[3,5]{Zhongji Zhang}[orcid=0009-0006-5454-6764]
\credit{Conceptualization, Methodology, Writing – original draft}

\author[3,5]{Yuhang Wang}[orcid=0009-0003-0141-3827]
\credit{Data curation, Software, Validation}

\author[3,5]{Yinghao Zhu}[orcid=0000-0002-2640-6477]
\credit{Data curation, Software, Writing - review \& editing}

\author[3,5]{Xinyu Ma}[orcid=0000-0003-4574-0830]
\credit{Data curation, Resources}

\author[3,5]{Yasha Wang}[orcid=0000-0002-8026-9688]
\credit{Formal analysis, Resources, Supervision}

\author[6,7]{Junyi Gao}[orcid=0000-0002-4951-8682]
\credit{Formal analysis, Resources, Supervision}

\author[3,5]{Liantao Ma}[orcid=0000-0001-5233-0624]
\credit{Resources, Supervision, Writing – review \& editing}

\author[4]{Wen Tang}
\credit{Formal analysis, Writing – review \& editing}

\author[2]{Xiaoyun Zhang}
\cormark[1]
\credit{Formal analysis, Writing – review \& editing}

\author[1]{Ling Wang}
\cormark[1]
\credit{Formal analysis, Writing – review \& editing}

\affiliation[1]{
organization={Affiliated Xuzhou Municipal Hospital of Xuzhou Medical University}, 
city={Jiangsu},
country={China}
}

\affiliation[2]{
organization={Peking University School and Hospital of Stomatology}, 
city={Beijing},
country={China}
}

\affiliation[3]{
organization={National Engineering Research Center for Software Engineering, Peking University},
city={Beijing},
country={China}
}

\affiliation[4]{
organization={Peking University Third Hospital}, 
city={Beijing},
country={China}
}

\affiliation[5]{
organization={Key Laboratory of High Confidence Software Technologies, Ministry of Education}, 
city={Beijing},
country={China}
}

\affiliation[6]{
organization={Centre for Medical Informatics, University of Edinburgh}, 
city={Edinburgh},
country={United Kingdom}
}

\affiliation[7]{
organization={Health Data Research UK}, 
country={United Kingdom}
}

\cortext[cor1]{Corresponding author.}

\begin{abstract}
Emerging diseases present challenges in symptom recognition and timely clinical intervention due to limited available information. An effective prognostic model could assist physicians in making accurate diagnoses and designing personalized treatment plans to prevent adverse outcomes. However, in the early stages of disease emergence, several factors hamper model development: limited data collection, insufficient clinical experience, and privacy and ethical concerns restrict data availability and complicate accurate label assignment. Furthermore, Electronic Medical Record (EMR) data from different diseases or sources often exhibit significant cross-dataset feature misalignment, severely impacting the effectiveness of deep learning models. We present a domain-invariant representation learning method that constructs a transition model between source and target datasets. By constraining the distribution shift of features generated across different domains, we capture domain-invariant features specifically relevant to downstream tasks, developing a unified domain-invariant encoder that achieves better feature representation across various task domains. Experimental results across multiple target tasks demonstrate that our proposed model surpasses competing baseline methods and achieves faster training convergence, particularly when working with limited data. Extensive experiments validate our method's effectiveness in providing more accurate predictions for emerging pandemics and other diseases. Code is publicly available at \url{https://github.com/wang1yuhang/domain_invariant_network}.
\end{abstract}



\begin{keywords}
domain-invariant \sep transfer learning \sep electronic medical record \sep emerging disease
\end{keywords}

\maketitle 

\section{Introduction}

Recent advancements in data analytics and machine learning are fundamentally transforming medical research and practice. Electronic Medical Records (EMR) have emerged as invaluable resources for generating clinical insights, enhancing prognostic models, enabling early diagnosis, and facilitating personalized treatment~\cite{zhu2024emerge,alam2020prediction,gao2020stagenet}.

The extensive, high-quality data accumulated through electronic medical information systems across healthcare institutions provides a robust foundation for accurate clinical predictions, including mortality risk assessment~\cite{ma2020adacare,zoref2022improved,zhu2024prism} and disease diagnosis~\cite{gao2019camp,ma2018health}. Various deep learning architectures~\cite{vaswani2017attention,dey2017gate,wang2020c} have demonstrated success in predicting diverse disease outcomes.

However, emerging diseases present a critical challenge: the initial scarcity of clinical experience and data significantly impedes the development of effective prognostic deep learning models~\cite{gao2020dr}. While predictive models require extensive labeled data for robust feature extraction~\cite{huang2020clinical}, data scarcity substantially compromises their performance and clinical utility~\cite{zhang2022m3care}. The applicability of existing models to new diseases is limited by multiple factors: insufficient data, minimal clinical experience, privacy restrictions, and ethical considerations. These constraints restrict access to reference data and hinder model effectiveness. Furthermore, the lack of comprehensive information complicates early-stage symptom recognition and outcome prediction, potentially delaying crucial clinical interventions and optimal medical resource allocation~\cite{weissman2020locally}.

Transfer learning has emerged as a promising approach for data-scarce medical scenarios, with numerous successful applications~\cite{ma2021distilling,ma2023aicare,mpau}. However, this approach encounters significant challenges when applied to emerging diseases, particularly due to cross-dataset feature misalignment that reduces model efficiency. The limited availability of shared features between source and target domains complicates knowledge transfer, emphasizing the need for robust, adaptable representation learning methods that maintain effectiveness across varying clinical domains.

Our research addresses the crucial challenge of developing a resilient, high-performance AI diagnostic assistance model using limited EMR data. We propose a new approach that leverages external datasets to train a versatile temporal EMR feature extractor. This method overcomes the limitations of insufficient and feature-misaligned data by learning domain-invariant representations for various target tasks, rather than relying on fixed information from the source domain.

Through comprehensive experimentation and comparative analysis against established baseline methods, we demonstrate our model's superior performance in both prediction accuracy and convergence speed, offering healthcare practitioners a reliable tool for predicting outcomes of emerging diseases.

Our primary contributions are:
\begin{itemize}
\item \textbf{Methodologically}, we introduce a domain-invariant representation learning method for prognosis prediction that effectively addresses data distribution shifts across EMR datasets, enabling accurate clinical prediction despite limited data and misaligned features. Our approach incorporates a Transition Model between source and target datasets, capturing domain-invariant features crucial for downstream tasks and facilitating the development of a unified domain-invariant encoder.

\item \textbf{Experimentally}, we present comprehensive experiments across multiple datasets. The results demonstrate consistent superior performance compared to baseline approaches, with notably higher training convergence rates, particularly in limited-data scenarios. Specifically, our solution achieves a 4.3\% reduction in MSE on the TJH dataset and an 8.5\% improvement when working with limited training samples, compared to the best-performing baseline methods.
\end{itemize}

\section{Related Work}

\subsection{Early-Stage Clinical Prediction for Emerging Diseases}

Early-stage clinical prediction for emerging diseases has become a critical research focus due to its potential to revolutionize healthcare interventions for novel and rapidly spreading illnesses. Accurate prediction of patient outcomes during the initial phases of emerging diseases is crucial for enabling timely clinical interventions, developing personalized treatment strategies, and optimizing medical resource allocation.

The widespread adoption of electronic medical information systems has led to the accumulation of extensive Electronic Medical Records (EMR) across healthcare institutions~\cite{feng2021completing}. These high-quality datasets have established a robust foundation for various clinical predictions, including mortality prediction~\cite{ma2020adacare,zoref2022improved} and diagnosis prediction~\cite{gao2019camp,ma2018health}. Numerous deep learning models~\cite{vaswani2017attention,dey2017gate} have been deployed to predict outcomes for various emerging diseases. Wang~\cite{wang2020c} and Liu et al.~\cite{liu2020neutrophil} investigated the correlations between early-stage biomarkers and disease severity in COVID-19 patients. Alam et al.~\cite{alam2020prediction} established prediction rules for scrub typhus meningoencephalitis, an emerging disease in North India, enabling physicians in peripheral areas to identify and treat this condition effectively. Gao et al.~\cite{gao2020stagenet} developed a method that leverages disease progression information from EMR for risk prediction tasks.

\subsection{Transfer Learning and Feature Alignment in Medical Contexts}

Transfer learning in medical applications has emerged as a promising approach to enhance predictive accuracy and clinical decision-making, particularly in scenarios with limited data availability. This methodology has demonstrated significant potential in addressing the challenges of data scarcity in healthcare settings.

Lopes et al. demonstrated the effectiveness of transfer learning by adapting parameters from a convolutional neural network pre-trained on a large dataset for sex detection. Although the initial task had limited clinical relevance, the model acquired valuable features that, when fine-tuned with just 310 recordings for heart disease detection, outperformed both models trained from scratch and clinical experts~\cite{lopes2021improving}.

In the realm of temporal medical data, Ma et al.~\cite{ma2021distilling} explored the application of TimeNet~\cite{malhotra2017timenet}, an unsupervised pre-trained model, for clinical feature extraction from non-medical time series. Their study revealed limitations in feature generalization, resulting in negative transfer effects. Wardi et al.~\cite{wardi2021predicting} demonstrated the superiority of transfer learning approaches over non-transfer methods in predicting septic shock using limited EMR data in emergency departments. Further advancing the field, Ma et al.~\cite{ma2021distilling} introduced a distilled transfer framework that leverages deep learning to embed features from extensive EMR data and transfers these parameters to a student model, which is subsequently trained to emulate the teacher model's representation, consistently outperforming baseline methods.

However, existing transfer learning approaches face significant challenges when applied to emerging diseases. First, the extreme scarcity of data in emerging disease scenarios poses a fundamental challenge. Second, when utilizing EMR datasets from different domains, the overlap in features between different diseases or prediction tasks is often minimal, representing only a small subset of all features. The selective transfer of these shared features results in substantial information loss from the source domain and creates inconsistent initialization levels between shared and private features during target domain model training, leading to learning bias and model deviation~\cite{seah2011healing,zhu2023m3fair}. Furthermore, to enhance clinical applicability for emerging diseases, models must develop the capability to learn domain-invariant representations adaptable to various target tasks, rather than merely acquiring fixed information from the source domain.

\section{Problem Formulation}

We formally define our research problem and establish the notation framework used throughout this paper, as summarized in Table~\ref{tab:notations}.

\begin{table}[!ht]
\centering
\caption{Notations used in the paper.}
\label{tab:notations}
\begin{tabular}{c|c}
\toprule
Notation & Definition\\ 
\midrule
$\bm{x}_{i}$ & Time-series record of the $i$-th medical feature\\
$\bm{f}_{i}$ & Feature embedding of the $i$-th medical feature\\
$\bm{s}$ & Overall representation of patient’s health status\\
$\bm{F}$ & Feature embedding matrix\\
\midrule
$\hat{\bm{y}}_{outcome}$ & Result of outcome prediction\\
$\hat{\bm{y}}_{LOS}$ & Result of LOS prediction\\
$\bm{y}_{outcome}$ & Groundtruth Label of outcome prediction\\
$\bm{y}_{LOS}$ & Groundtruth Label of LOS prediction\\
\midrule
$n_{sf}$ & Number of shared features\\
$n_{pf, src}$ & Number of private features in source dataset\\
$n_{pf, tar}$ & Number of private features in target dataset\\
\bottomrule
\end{tabular}
\end{table}

\textbf{Electronic Medical Record (EMR) Representation}: Electronic Medical Records comprise longitudinal patient observations collected during clinical admissions. These observations consist of both static baseline information (e.g., age, gender, primary diagnosis) and dynamic time-series features (e.g., medications, diagnoses, vital signs, laboratory measurements). For each clinical admission, we observe $N$ distinct features, where each feature $\bm{x}_{i} \in \mathbb{R}^{T}(i = 1, 2, \cdots, N)$ represents a time series of measurements. Each medical feature contains $T$ sequential timestamps. These clinical sequences are organized into a longitudinal patient matrix $\bm{x}$, where rows correspond to medical features and columns represent temporal measurements.

\textbf{Clinical Prediction Tasks}: In the context of emerging diseases, early assessment of patient health trajectories is crucial for clinical decision-making. Given that patients may experience varying outcomes and resource utilization patterns based on disease severity~\cite{gao2024comprehensive}, we formulate a dual-objective prediction task that simultaneously addresses two critical clinical outcomes:

\begin{enumerate}
    \item \textbf{Mortality Prediction}: Given the longitudinal patient matrix $\bm{x}$, predict the binary outcome ${\bm{y}}_{outcome} \in \{0, 1\}$, where ${\bm{y}}_{outcome} = 1$ indicates mortality and ${\bm{y}}_{outcome} = 0$ denotes survival.
    \item \textbf{Length-of-Stay Prediction}: Estimate ${\bm{y}}_{LOS}$, representing the duration of the patient's healthcare facility stay, based on the same input matrix $\bm{x}$.
\end{enumerate}

This dual-objective formulation is particularly significant for emerging diseases as it enables physicians to assess disease severity, identify critical cases requiring immediate intervention, facilitate novel treatment strategies, support evidence-based policy formulation, and optimize resource allocation during healthcare system strain.

In addition, the remaining life span prediction task involves estimating the duration between initial diagnosis and mortality. This regression task is particularly challenging due to the complex progression patterns and the numerous factors affecting patient survival. The prediction target $\bm{y}_{lifespan}$ represents the number of days a patient survives after diagnosis. This task differs from traditional mortality prediction as it requires the model the temporal progression of the disease, making it a more comprehensive evaluation of model performance in handling complex clinical trajectories.

\section{Methodology}

\subsection{Overview}

\begin{figure}[!ht]
\centering
\includegraphics[width=0.9\linewidth]{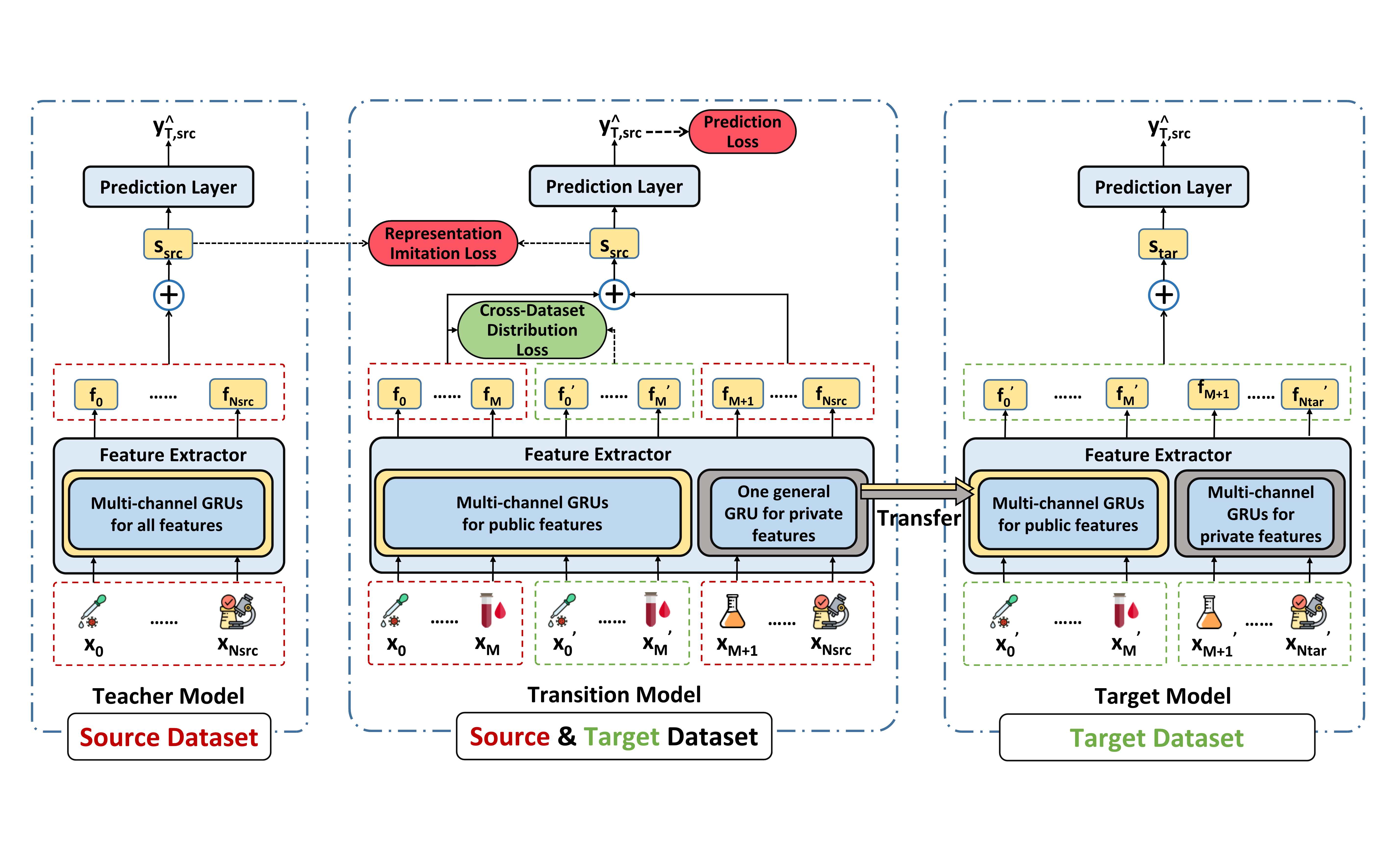}
\caption{The overall pipeline of the proposed method. Left: Training teacher model on source dataset. Middle: Training domain-invariant feature extractor. Right: Transferring to target task. (Features and vectors generated from \textcolor{red}{source dataset} are colored in \textcolor{red}{RED}. Features and vectors generated from \textcolor{green}{target dataset} are colored in \textcolor{green}{GREEN}).}
\label{fig:framework}
\end{figure}

To address the challenges of limited data availability and complex clinical prediction tasks in emerging diseases, we propose a domain-invariant representation learning method that leverages existing large-scale EMR source datasets and their corresponding models. The overall pipeline is shown in Figure~\ref{fig:framework}. The wide variety of medical features and varying clinical feature requirements across different tasks necessitate a strategic approach to mitigate distribution shifts between domains. Building upon a pre-trained teacher model, our solution involves training a transitional model on the source dataset while incorporating information from the target dataset. By combining knowledge guidance from the teacher model and supervision from the target dataset, our approach focuses on capturing domain-invariant features relevant to downstream tasks, facilitating the learning of a unified domain-invariant encoder. The primary goal is to enhance the model's generalization capability across different task domains, ultimately improving its adaptability to various medical contexts. The implementation consists of the following three key steps: (1) training teacher model on source dataset, (2) training domain-invariant feature extractor, and (3) transferring to target task.

\subsection{Training Teacher Model on Source Dataset}

The initial step involves training a teacher model on a large-scale EMR dataset to establish a strong supervisory signal for the transitional model training. The teacher model serves as a guiding entity during the transfer learning process. Research has shown that GRU often outperforms other RNN structures on EMR datasets~\cite{ma2020concare,ma2023aicare}. To leverage the advantages demonstrated by multi-channel GRU in capturing patient health state representations, we adopt a multi-channel GRU architecture to capture distinctive patterns from different medical features. We create N separate GRUs, where each GRU is responsible for embedding one dynamic medical feature. For each dynamic feature $i$, we represent it as a time series $\bm{x}_i = (x_{i1}, x_{i2}, \cdots, x_{iT})$, which is fed into the corresponding GRU$_{i}$ to generate feature embedding: 
\begin{equation}
\bm{f}_{i} = \mathrm{GRU}_{i}(x_{i1}, x_{i2}, \cdots, x_{iT}). 
\end{equation}

The resulting embedding matrices $\bm{f}_{i}$ are stacked to create the feature embedding matrix $\mathbf{F} = (\bm{f}_{0}, \bm{f}_{1},\bm{f}_{2}, \cdots, \bm{f}_{N}) ^ {\mathsf{T}}$. This matrix is then input into a linear layer to extract the patient's health representation $\bm{s}_{src}$, which is used to obtain the final prediction of source dataset $\hat{\bm{y}}_{src} \in \mathbf{R}$:
\begin{equation}
\bm{s}_{src} = \bm{F} \cdot \mathbf{W}_1 + \bm{b}_1
\end{equation}
\begin{equation}
\hat{\bm{y}}_{src} = \mathrm{Prediction Layer}({\bm{s}}_{src})
\end{equation}

For regression tasks, we adopt mean square error (MSE) as the loss of source dataset $\mathcal{L}_{pred}$:
\begin{equation}
\mathcal{L}_{pred} = \frac{1}{n}\sum_{i=1}^{n} (y{_{src}^{(i)}} - \hat{y}{_{src}^{(i)}}) ^2
\end{equation}

For binary classification with label ${\bm{y}}_{T, src}$, we apply sigmoid activation and cross-entropy loss for computing the prediction loss $\mathcal{L}_{pred}$.

\subsection{Training Domain-Invariant Feature Extractor}

While the feature extractor of the teacher model excels at capturing patients' health state representation within the source dataset, it often struggles with diverse target datasets and prediction tasks due to task variations, potentially leading to negative transfer. To address this, we mitigate the distribution shift of generated features between different domains by capturing domain-invariant features relevant to downstream tasks.

Given feature misalignment in EMR data, the source and target datasets may possess both shared features $\bm{x}_{sf} = (\bm{x}_{1}, \bm{x}_{2}, \cdots, \bm{x}_{M})$ and unique private features $\bm{x}_{pf, src} = (\bm{x}_{1}, \bm{x}_{2}, \cdots, \bm{x}_{N_{src}})$, $\bm{x}_{pf, tar} = (\bm{x}_{1}, \bm{x}_{2}, \cdots, \bm{x}_{N_{tar}})$, where $M$ is the number of shared features, $N_{src}$ and $N_{tar}$ represent the feature counts in the source and target datasets, respectively. The multi-channel GRU feature extractor aligns features, restricting domain transfer predominantly to shared features. This hampers both model generalization and utilization of essential information in private features. Our transitional model must therefore handle both shared and private features effectively.

For shared features across source and target datasets, the feature extraction layer of the transitional model mirrors the teacher model design, with each shared feature corresponding to a GRU. To promote domain-invariant feature learning relevant to downstream tasks, we ensure samples from different domains exhibit similar hidden-space distributions after encoding. The shared features of source $\bm{x}_{sf, src}$ and target $\bm{x}_{sf, tar}$ are simultaneously input into the multi-channel GRU of the transitional model, generating feature embedding matrices:
\begin{equation}
\bm{F}_{sf, src}, \bm{F}_{sf, tar} = \mathrm{MCGRU}(\bm{x}_{sf, src}, \bm{x}_{sf, tar})
\end{equation}

We employ adversarial learning to achieve domain-invariant representations between source and target domains. A well-trained domain classifier D is introduced, and the feature extractor should generate embeddings that confound this classifier. The multi-channel GRUs are adversarially updated through a gradient reversal layer, ensuring correct prediction layer performance while confusing the domain classifier about the source of feature embeddings.

Given a domain classifier D with parameters $\theta_d$, multi-channel GRU with parameters $\theta_f$, and prediction layer with parameters $\theta_p$, the adversarial loss is:
\begin{equation}
\begin{split}
    \arg\min_{(\theta_f, \theta_p)}( L_{pred} - L_d(D(F_{sf, src}), d_{S})
    \\ - L_d(D(F_{sf, tar}),d_{T}))
\end{split}
\end{equation}
where $L_d,L_{pred}$ are multi-class cross-entropy losses for domain classification and prediction tasks, and $d_S, d_T$ are domain labels for source and target data.

Using feature embedding matrices of shared features $\bm{F}_{sf, src}$, we obtain complete feature embedding matrices of source datasets $\bm{F}_{src}$ and compute the patient's health representation $\bm{s}_{src}$:
\begin{equation}
\bm{s}_{src} = \bm{F}_{src} \cdot \mathbf{W}_1 + \bm{b}_1
\end{equation}

The representation $\bm{s}_{src}$ should imitate $\tilde{\bm{s}}_{src}$ generated by the teacher model. The representation simulation loss $\mathcal{L}_{rep}$ is defined using KL-Divergence: 
\begin{equation}
\mathcal{L}_{rep} = \mathrm{D_{KL}}(\mathrm{Softmax}(\tilde{\bm{s}}_{src}) || \mathrm{Softmax}(\bm{s}_{src})) 
\end{equation}
\begin{equation}
\mathrm{D_{KL}}(P||Q) = \sum_{i} P_i \log(\frac{P_i}{Q_i})
\end{equation}

Teacher model supervision enables the transitional model to mimic its behavior, aiding in comprehensive feature correlation capture and holistic health state representation generation.

While capturing domain-invariant features, the transitional model must not disclose target dataset-specific prediction information. Therefore, it only makes predictions on the source dataset, consistent with the teacher model. The total loss $\mathcal{L}$ combines three components:
\begin{equation}
\mathcal{L} = \alpha\mathcal{L}_{rep} + \beta\mathcal{L}_{pred} -\gamma\mathcal{L}_{d}
\end{equation}
where $\alpha$, $\beta$, $\gamma$ are hyperparameters balancing these losses, all set to 1 in our experiments.

\subsection{Transferring to Target Task}

Finally, we transfer GRUs from the transitional model to the target model and fine-tune on the target dataset. The target model structure mirrors the teacher model, using multi-channel GRUs for feature extraction. For shared features, we transfer the corresponding GRUs from the transitional model.

\begin{figure}[!ht]
\centering
\includegraphics[width=0.4\linewidth]{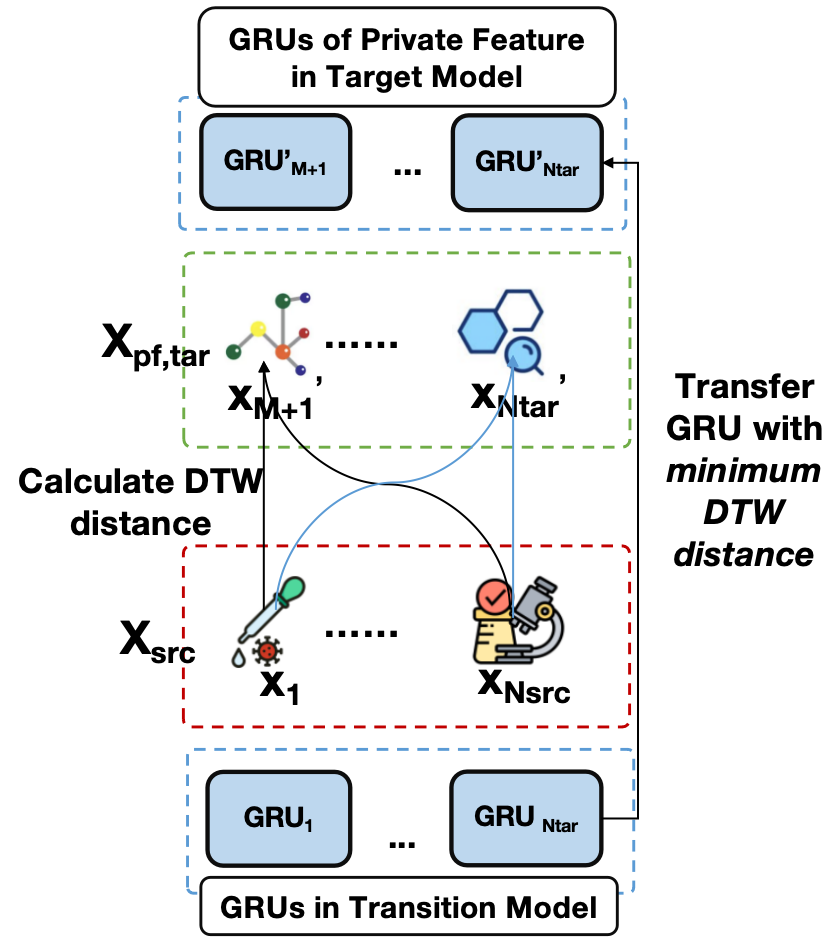}
\caption{Transfer GRUs to target model based on DTW distance between features.}
\label{fig:DTW}
\end{figure}

For private features unique to the target dataset, we employ a transfer approach based on feature embedding distance, as shown in Figure~\ref{fig:DTW}. Dynamic Time Warping (DTW)~\cite{muller2007dynamic} is chosen as the similarity measure due to its capability to handle time series data with varying time steps, without considering semantic feature meanings~\cite{fawaz2018transfer}. During target model initialization, for each private target feature, we calculate its DTW distance with each source feature, selecting the source feature with the smallest DTW value as most similar in sequential information content. The corresponding GRU parameters are then transferred, ensuring consistent initialization levels across all target model features.

The target model employs two separate MLP prediction heads for patient mortality $\hat{\bm{y}}_{tar, outcome}$ and length-of-stay (LOS) $\hat{\bm{y}}_{tar, LOS}$ predictions. Mean squared error (MSE) loss is used for LOS prediction, while binary cross-entropy (BCE) loss is used for mortality prediction.

\begin{algorithm}[!ht]
\caption{Algorithm}
\label{alg:full_model}
\begin{algorithmic}[1] 
\STATE \textbf{Stage 1:} Training teacher model on source dataset
\STATE Let parameters in teacher model randomly initialized.
\WHILE{not convergence}
\STATE Compute $\hat{\bm{y}}_{src}, \tilde{\bm{s}}_{src}$
\STATE Compute $\mathcal{L}_{pred} = \mathrm{MSE}(\hat{\bm{y}}_{src}, \bm{y}_{src})$
\STATE Update parameters of Teacher Model by optimizing $\mathcal{L}_{pred}$ using back-propagation.
\ENDWHILE
\STATE \textbf{Stage 2:} Training domain-invariant feature extractor
\STATE Let parameters in transition model randomly initialized.
\WHILE{not convergence}
\STATE Compute $\bm{F}_{sf, src}, \bm{F}_{sf, tar}, \bm{F}_{pf, src}, \hat{\bm{y}}_{src}, {\bm{s}}_{src}$
\STATE Compute $\mathcal{L}_{pred}, \mathcal{L}_{sup}, \mathcal{L}_{d}$
\STATE Update parameters of teacher model by optimizing $\alpha\mathcal{L}_{sup} + \beta\mathcal{L}_{pred} - \gamma\mathcal{L}_{d}$ using back-propagation.
\ENDWHILE
\STATE \textbf{Stage 3:} Transferring to target task
\STATE Transfer parameters of shared GRUs to target model's GRUs for shared features, and parameters of private GRUs to target model's GRUs for private features based on DTW distance.
\WHILE{not convergence}
\STATE Compute $\hat{\bm{y}}_{tar, LOS}, \hat{\bm{y}}_{tar, outcome}$
\STATE Compute $\mathcal{L}_{tar} = \mathrm{BCE}(\hat{\bm{y}}_{tar, outcome}, {\bm{y}}_{tar, outcome}) + \mathrm{MSE}(\hat{\bm{y}}_{tar, LOS}, {\bm{y}}_{tar, LOS})$
\STATE Update parameters of target model by optimizing $\mathcal{L}_{tar}$ using back-propagation.
\ENDWHILE
\end{algorithmic}
\end{algorithm}

The specific process of the algorithm is shown in Algorithm~\ref{alg:full_model}.

\section{Experiments}

We evaluate our proposed method on multiple real-world EMR datasets to demonstrate its effectiveness in clinical prediction tasks. The PhysioNet dataset serves as our source dataset for enhancing target task predictions. To verify our model's scalability across different clinical prediction tasks and EMR datasets, we employ three target datasets: two COVID-19 datasets for the multitask prediction setting, and one end-stage renal disease (ESRD) dataset for remaining life span prediction. Table~\ref{tab:statistics} presents comprehensive statistics of these datasets. We additionally evaluate performance using reduced training samples from the TJH dataset to simulate early-stage emerging disease scenarios with limited data.

\subsection{Dataset Description}

\subsubsection{Sepsis Source Dataset from PhysioNet Dataset}

The PhysioNet dataset~\cite{reyna2019early}, chosen for its comprehensive nature and widespread use in the field, serves as our source dataset for pre-training the teacher model on sepsis prediction. This dataset comprises ICU patient records from three hospitals collected over the past decade. It contains 34 distinct clinical variables, including 8 vital signs and 26 laboratory measurements, with hourly summaries of vital signs (e.g., heart rate, pulse oximetry) and laboratory measurements (e.g., creatinine, calcium).

\subsubsection{Target Datasets}

\textbf{Tongji Hospital (TJH) COVID-19 Dataset:} This dataset~\cite{yan2020interpretable} contains medical records of 361 COVID-19 patients from Tongji Hospital, spanning January 10th to February 18th, 2020. The dataset includes 195 recovered patients and 166 deceased patients.

\textbf{HM Hospitals (HMH) COVID-19 Dataset:} The HMH dataset~\cite{hmh} includes anonymous records of 4,255 confirmed or suspected COVID-19 patients with at least one laboratory test record. Of these patients, 540 did not survive.

\textbf{Peking University Third Hospital (PUTH) ESRD Dataset:} This dataset comprises records of 325 ESRD patients who received treatment between January 1st, 2006, and March 1st, 2018. The remaining life span prediction task for this dataset involves estimating the survival duration of patients from their initial diagnosis.

\begin{table}[!ht] 
\footnotesize
\centering
\caption{Statistics of datasets used.}
\label{tab:statistics}
\begin{tabular}{c|c|ccc} 
\toprule
& Source & \multicolumn{3}{c}{Target}\\
Dataset & PhysioNet & TJH & HMH & ESRD\\
\midrule
\#Patients & 40336 & 361 & 4255 & 325\\
\#Records & 1552210 & 1704 & 123044 & 10787\\
\#Features & 34 & 75 & 99 & 69\\
\#Shared Features & / & 18 & 26 & 16\\
\bottomrule
\end{tabular}
\end{table}

As shown in Table~\ref{tab:statistics}, substantial feature overlap exists between source and target datasets, with shared features comprising 24\%, 26\%, and 23\% of total features for TJH, HMH, and PUTH datasets respectively. Table~\ref{tab:shared_features} details these shared features, which provide valuable reference information for knowledge transfer.

\begin{table}[!ht]
\footnotesize
\centering
\caption{Shared features across source and target datasets.} 
\label{tab:shared_features}
\begin{tabular}{c|c|c} 
\toprule
PhysioNet-TJH & PhysioNet-HMH & PhysioNet-ESRD\\
\midrule
HCO$_{3}^{-}$ & Heart rate & Heart rate\\
PH value & Pulse oximetry & Systolic BP\\
Urea & Temperature & Diastolic BP\\
Alkalinephos & Systolic BP & EtCO$_{2}$\\
Calcium & Diastolic BP & Urea\\
Chloride & HCO$_{3}^{-}$ & Alkalinephos\\
Creatinine & PH value & Calcium\\
Bilirubin direct & PaCO$_{2}$ & Chloride\\
Glucose & SaO$_{2}$ & Creatinine\\
Potassium & Calcium & Glucose\\
Bilirubin total & Chloride & Phosphate\\
Troponin I & Creatinine & Potassium\\
Hematocrit & Bilirubin direct & Hct\\
Hemoglobin & Glucose & Hgb\\
aPTT & Lactate & WBC\\
WBC & Magnesium & Platelets\\
Fibrinogen & Phosphate & /\\
Platelets & Potassium & /\\
/ & Bilirubin total & /\\
/ & Troponin I & /\\
/ & Hct & /\\
/ & Hgb & /\\
/ & aPTT & /\\
/ & WBC & /\\
/ & Fibrinogen & /\\
/ & Platelets & /\\
\bottomrule
\end{tabular}
\end{table}

\subsection{Experimental Setups}

We implement rigorous cross-validation procedures to prevent data leakage, ensuring patient records remain isolated across different folds. Our preprocessing methods and prediction task selections align with the benchmark established by~\cite{gao2024comprehensive}. The PUTH dataset is preprocessed following the protocol~\cite{wang2024protocol}. All experiments are conducted on a server equipped with an Nvidia RTX 3090 GPU and 64GB RAM, using CUDA 11.2, Python 3.7, and PyTorch 1.12.1.

We employ 5-fold cross-validation for all prediction tasks. For regression tasks, we use Mean Square Error (MSE) and Mean Absolute Error (MAE)~\cite{chai2014root} as evaluation metrics. For classification tasks, we utilize the Area Under Receiver Operating Characteristic Curve (AUROC)~\cite{fawcett2006introduction}. All baseline models are implemented using the pyehr package~\cite{zhu2023pyehr}.

The baseline methods include:

\begin{itemize}

\item \textit{GRU}~\cite{dey2017gate} is a type of \textit{Recurrent Neural Network (RNN)} architecture. 
\item \textit{Transformer}~\cite{vaswani2017attention} leverage its self-attention mechanism to enhance its ability in capturing long-range dependencies of time-series data.
\item \textit{T-LSTM}~\cite{baytas2017patient} is a time-aware network which can handle irregular time intervals in longitudinal records and learn fixed-dimensional representations.
\item \textit{Concare}~\cite{ma2020concare} is a powerful deep learning method which models the static and dynamic data by embedding the features separately and using the self-attention mechanism.
\item \textit{StageNet}~\cite{gao2020stagenet} is a powerful deep learning method which extracts the information of different stage of diseases from the EMR and then utilize it in the risk prediction task.
\item \textit{TimeNet}~\cite{malhotra2017timenet} is an innovativate transfer learning method for learning features from time series data, which can maps varying length and complex time series data from different domains.
\item \textit{DistCare}~\cite{ma2021distilling} is a distilled transfer learning framework that uses deep learning to transfer knowledge from online EMR data to improve the prognosis of patients with new diseases.

\item \textit{Dann}~\cite{ganin2015unsupervised} a classic domain adaptation framework, which was one of the earliest works to introduce adversarial learning frameworks into transfer learning

\item \textit{Codats}~\cite{wilson2020multi} is a domain adaptation framework which based on CNN-1D for time series data using adversal learning.

\end{itemize}

\subsection{Experimental Results}

\subsubsection{Benchmarking Performance}

\begin{table}[!ht]
\footnotesize
\centering
\caption{ESRD prognosis prediction performance on the PUTH dataset (remaing life span).}
\label{tab:result_ESRD}
\begin{tabular}{c|cc} 
\toprule
& \multicolumn{2}{c}{PUTH dataset}\\
Methods & MSE & MAD\\
\midrule
GRU & 700.375(73.017) & 21.710(0.948)\\
Transformer & 666.060(87.263) & 20.290(1.658)\\
Concare & 646.039(79.037) & 20.418(1.268)\\
StageNet& 659.011(30.928) & 21.579(0.436)\\
TimeNet & 690.704(12.840) & 21.232(0.229)\\
T-LSTM & 640.689(89.942) & 20.419(1.517)\\
DistCare & 632.932(74.644) & 20.723(1.420)\\
Dann & 633.051(60.905) & 20.072(0.999) \\
Codats & 648.445(92.112) & 20.367(1.585)\\
\midrule
$Ours_{tar}$ & 621.179(78.063) & 20.062(1.204)\\
$Ours$ & \textbf{610.231}(67.424) & \textbf{19.987}(1.058)\\
\bottomrule
\end{tabular}
\end{table}

\begin{figure}[!ht]
\centering
\includegraphics[width=0.5\linewidth]{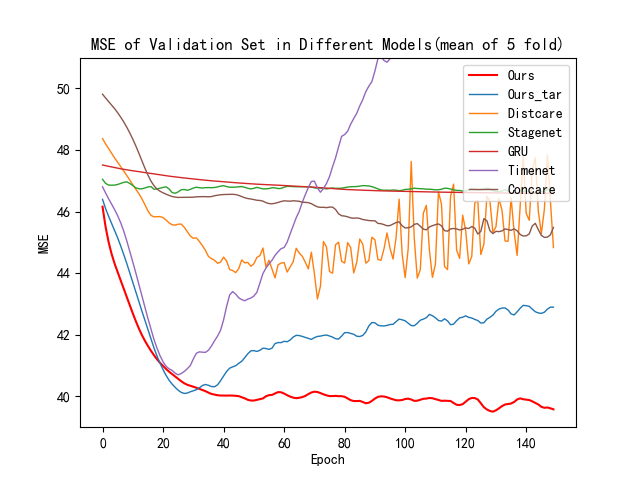} 
\caption{Comparison of the convergence speeds of different models on the TJH dataset (Our model is indicated by the bold \textcolor{red}{RED} line).}
\label{fig:MSE}
\end{figure}

The experimental results are shown in Table~\ref{tab:result_ESRD} and Table~\ref{tab:result_TJH_HMH}. The $Ours_{tar}$ in these table is the performance of our non-transfered target model as ablation experiment results. Our solution consistently outperforms both transfer-based and non-transfer-based baselines, which demonstrating the advantages of our model in predicting emerging diseases. We also showcase a comparison of the convergence speeds of different models, which is shown in Figure~\ref{fig:MSE}. Each curve represents a model's average validation MSE values across five-fold cross-validation as a function of training epochs. Our model demonstrates a pronounced convergence speed advantage in the early stages, rapidly achieving lower validation MSE values. Comparing to other models, our approach attains higher accuracy within a limited number of training iterations. This outcome underscores the efficiency and superiority of our model's learning process, providing fast and accurate support for clinical prediction of emerging diseases.

\subsubsection{Performance under Limited Data Scenarios}

\setlength{\tabcolsep}{3pt}
\begin{table*}[!ht]
\scriptsize
\centering
\caption{COVID-19 prognosis prediction performance on the TJH and HMH datasets (multitask prediction).}
\label{tab:result_TJH_HMH}
\begin{tabular}{c|ccc|ccc}
\toprule
& \multicolumn{3}{c|}{TJH dataset} & \multicolumn{3}{c}{HMH dataset}\\
Methods & MSE & MAD & AUROC & MSE & MAD & AUROC\\
\midrule
GRU & 46.522(4.601) & 5.327(0.153) & 0.509(0.003) & 35.096(2.000) & 4.305(0.024) & 0.503(0.0003)\\
Transformer & 45.079(7.362) & 5.013(0.310) & 0.942(0.018) & 31.964(1.439) & 3.933(0.039) & 0.835(0.017)\\
Concare  & 41.835(8.599) & 5.082(0.430) & 0.866(0.151) & 31.736(2.005) & 4.245(0.048) & 0.800(0.007)\\
StageNet & 46.486(4.838) & 5.319(0.180) & 0.957(0.019) & 34.981(1.968) & 4.257(0.023) & 0.777(0.012)\\
TimeNet & 38.310(4.088) & 4.576(0.136) & \textbf{0.986}(0.006) & 28.714(1.870) & 3.812(0.079) & 0.890(0.005)\\
T-LSTM  & 37.684(5.464) & 4.388(0.316) & 0.957(0.011) & 29.121(2.294) & 3.747(0.084) & 0.841(0.020)\\
DistCare & 37.028(6.187) & 4.433(0.292) & 0.976(0.006) & 28.453(1.987) & 3.789(0.091) & 0.862(0.007)\\
Dann & 35.620(8.026) & 4.270(0.467) & 0.981(0.016) & 28.302(2.464) & 3.767(0.102) & 0.882(0.007) \\
Codats & 41.625(8.547) & 4.875(0.477) & 0.725(0.092) & 32.507(1.880) & 4.147(0.082) & 0.674(0.063) \\
\midrule
$Ours_{tar}$ & 36.415(3.647) & 4.403(0.156) & 0.982(0.006) & 28.787(1.963) & 3.824(0.035) & \textbf{0.890}(0.009)\\
$Ours$ & \textbf{35.432}(4.149) & \textbf{4.297}(0.177) & \textbf{0.983}(0.006)& \textbf{28.122}(2.141) & \textbf{3.782}(0.100) & \textbf{0.871}(0.002)\\
\bottomrule
\end{tabular}
\end{table*}

\setlength{\tabcolsep}{3pt}
\begin{table*}[t]
\scriptsize
\centering
\caption{COVID-19 prognosis prediction performance with fewer training samples on the TJH and HMH datasets (multitask prediction).}
\label{tab:result_TJH_HMH_reverse}
\begin{tabular}{c|ccc|ccc}
\toprule
& \multicolumn{3}{c|}{TJH dataset} & \multicolumn{3}{c}{HMH dataset}\\
Methods & MSE & MAD & AUROC & MSE & MAD & AUROC\\
\midrule
GRU & 46.570(1.205) & 5.326(0.025) & 0.506(0.001) & 35.108(0.994) & 4.296(0.012) & 0.501(0.001)\\
Transformer & 53.455(3.857) & 5.449(0.167) & 0.932(0.020) & 32.871(0.979) & 3.966(0.015) & 0.813(0.006)\\
Concare & 45.005(1.195) & 5.289(0.054) & 0.828(0.039) & 32.056(0.853) & 4.252(0.040) & 0.742(0.007)\\
StageNet & 46.273(1.157) & 5.281(0.020) & 0.961(0.007) & 35.021(1.018) & 4.261(0.030) & 0.798(0.044)\\
TimeNet & 40.474(1.553) & 4.685(0.027) & 0.969(0.010) & 30.478(1.061) & 3.978(0.046) & 0.852(0.002)\\
T-LSTM & 42.612(2.457) & 5.014(0.111) & 0.933(0.035) & 30.584(1.498) & 3.896(0.096) & 0.848(0.011)\\
DistCare & 42.844(1.146) & 4.971(0.240) & 0.818(0.067) & 29.672(0.532) & 3.865(0.040) & 0.843(0.013)\\
Dann & 40.151(1.403) & 4.597(0.112) & 0.951(0.020) & 29.932(0.725) & 3.853(0.011) & 0.859(0.017)\\
Codats & 45.063(3.17) & 5.173(0.294) & 0.592(0.121) & 32.540(0.606) & 4.187(0.042) & 0.670(0.027)\\
\midrule
$Ours$ & \textbf{39.215}(1.582) & \textbf{4.586}(0.177) & \textbf{0.976}(0.004) & \textbf{29.339}(0.823) & \textbf{3.858}(0.018) & \textbf{0.852}(0.006)\\
\bottomrule
\end{tabular}
\end{table*}

Furthermore, to simulate the circumstance of limited data and inadequate clinical information during the early stages of an emerging disease outbreak, we further reverse the training and testing sets of TJH and HMH dataset, making training samples fewer to evaluate performance. The results of this experiment are shown in Table~\ref{tab:result_TJH_HMH_reverse}.

Compared to other transfer learning baselines, our method shows improvements in most metrics on both datasets. For example, relative to Dann, Distcare, timenet and Codats,  our method achieves {2.33\%$\downarrow$}, {8.47\%$\downarrow$}, {3.11\%$\downarrow$} and {12.97\%$\downarrow$} MSE respectively on the TJ dataset. Similarly, on the HM dataset, our method achieves {1.98\%$\downarrow$}, {1.12\%$\downarrow$}, {3.74\%$\downarrow$} and  {9.83\%$\downarrow$} MSE.

\subsubsection{Ablation Study}
To explore the impact of different components on the experimental results on TJH dataset, we designed a comprehensive ablation experiment. The detailed results of these ablation studies are provided in Table~\ref{tab:ablation}, where $Ours_{tar}$ represents the performance of non-transfered target model, $Ours_{share}$ represents the performance of target model where only shared features are transferred and $Ours_{n\_tea}$ represents the performance of target model without training teacher model.

\begin{table}[!ht]
\footnotesize
\centering
\caption{Results of ablation study on TJH dataset.}
\label{tab:ablation}
\begin{tabular}{{c|c|c|c}}
\toprule
Baseline & MSE & MAE & Auroc \\
\midrule
$Ours_{tar}$ & 36.415(3.647) & 4.403(0.156) & 0.982(0.006)\\
$Ours_{share}$ & 35.886(7.790) & 4.386(0.47) & 0.983(0.017)\\
$Ours_{n\_tea}$ & 35.483(9.534) & 4.311(0.584) & 0.971(0.018)\\
\midrule
$Ours$ & \textbf{35.432}(4.149) & \textbf{4.297}(0.177) & \textbf{0.983}(0.006)\\
\bottomrule
\end{tabular}
\end{table}

The ablation results demonstrate the effectiveness of our model's key components. Compared to $Ours_{tar}$, our full model achieves a 2.7\% reduction in MSE and a 2.4\% reduction in MAE, highlighting the benefits of our transfer learning approach. The performance gap between $Ours_{share}$ and our full model (1.3\% reduction in MSE) validates the importance of transferring knowledge from both shared and private features. Additionally, the comparison with $Ours_{n\_tea}$ shows that the teacher model guidance improves prediction stability, as evidenced by the smaller standard deviations in our full model's results (4.149 vs 9.534 for MSE). While $Ours_{n\_tea}$ achieves competitive performance in terms of absolute metrics, the higher variance in its predictions suggests less reliable generalization. These results collectively demonstrate that each component of our proposed method contributes to its overall superior performance and stability.

\section{Conclusion}

This paper presents a new domain-invariant learning method for diverse clinical prediction tasks. Our approach effectively addresses the critical challenges of feature misalignment in EMR datasets and bridges data distribution shifts through a teacher model-guided learning framework supervised by target dataset objectives. Experimental results across multiple real-world EMR datasets demonstrate consistent superior performance compared to state-of-the-art baselines, particularly in limited-data scenarios typical of emerging diseases.

The success of our method lies in its ability to learn domain-invariant representations while maintaining high prediction accuracy across different clinical tasks. This capability is especially valuable for emerging disease scenarios where data scarcity poses significant challenges to traditional approaches. Our solution provides a robust framework for intelligent prognosis in future emerging disease scenarios, potentially enabling earlier and more accurate clinical interventions.

Future work could explore the integration of additional domain knowledge and the extension of our framework to other clinical prediction tasks. The methodology presented here represents a significant step forward in addressing the challenges of medical AI deployment in resource-constrained and emerging disease scenarios.



\printcredits

\section{Acknowledgments}

This work was supported by the National Natural Science Foundation of China (62402017, U23A20468), Beijing Natural Science Foundation (L244063), Xuzhou Scientific Technological Projects (KC23143), Peking University Medicine plus X Pilot Program-Key Technologies R\&D Project (2024YXXLHGG007). Junyi Gao acknowledges the receipt of studentship awards from the Health Data Research UK-The Alan Turing Institute Wellcome PhD Programme in Health Data Science (Grant Ref: 218529/Z/19/Z).

\bibliographystyle{unsrt}

\bibliography{ref}

\begin{thebibliography}{10}

\bibitem{zhu2024emerge}
Yinghao Zhu, Changyu Ren, Zixiang Wang, Xiaochen Zheng, Shiyun Xie, Junlan Feng, Xi~Zhu, Zhoujun Li, Liantao Ma, and Chengwei Pan.
\newblock Emerge: Enhancing multimodal electronic health records predictive modeling with retrieval-augmented generation.
\newblock In {\em Proceedings of the 33rd ACM International Conference on Information and Knowledge Management}, pages 3549--3559, 2024.

\bibitem{alam2020prediction}
Areesha Alam, Pranshi Agarwal, Jayanti Prabha, Amita Jain, Raj~Kumar Kalyan, Chandrakanta Kumar, and Rashmi Kumar.
\newblock Prediction rule for scrub typhus meningoencephalitis in children: emerging disease in north india.
\newblock {\em Journal of child neurology}, 35(12):820--827, 2020.

\bibitem{gao2020stagenet}
Junyi Gao, Cao Xiao, Yasha Wang, Wen Tang, Lucas~M Glass, and Jimeng Sun.
\newblock Stagenet: Stage-aware neural networks for health risk prediction.
\newblock In {\em Proceedings of The Web Conference 2020}, pages 530--540, 2020.

\bibitem{ma2020adacare}
Liantao Ma, Junyi Gao, Yasha Wang, Chaohe Zhang, Jiangtao Wang, Wenjie Ruan, Wen Tang, Xin Gao, and Xinyu Ma.
\newblock Adacare: Explainable clinical health status representation learning via scale-adaptive feature extraction and recalibration.
\newblock {\em Proceedings of the AAAI Conference on Artificial Intelligence}, 34(01):825--832, Apr. 2020.

\bibitem{zoref2022improved}
Adi Zoref-Lorenz, Jun Murakami, Liron Hofstetter, Swaminathan Iyer, Ahmad~S Alotaibi, Shehab~Fareed Mohamed, Peter~G Miller, Elad Guber, Shiri Weinstein, Joanne Yacobovich, et~al.
\newblock An improved index for diagnosis and mortality prediction in malignancy-associated hemophagocytic lymphohistiocytosis.
\newblock {\em Blood, The Journal of the American Society of Hematology}, 139(7):1098--1110, 2022.

\bibitem{zhu2024prism}
Yinghao Zhu, Zixiang Wang, Long He, Shiyun Xie, Xiaochen Zheng, Liantao Ma, and Chengwei Pan.
\newblock Prism: Mitigating ehr data sparsity via learning from missing feature calibrated prototype patient representations.
\newblock In {\em Proceedings of the 33rd ACM International Conference on Information and Knowledge Management}, pages 3560--3569, 2024.

\bibitem{gao2019camp}
Jingyue Gao, Xiting Wang, Yasha Wang, Zhao Yang, Junyi Gao, Jiangtao Wang, Wen Tang, and Xing Xie.
\newblock Camp: Co-attention memory networks for diagnosis prediction in healthcare.
\newblock In {\em 2019 IEEE international conference on data mining (ICDM)}, pages 1036--1041. IEEE, 2019.

\bibitem{ma2018health}
Tengfei Ma, Cao Xiao, and Fei Wang.
\newblock Health-atm: A deep architecture for multifaceted patient health record representation and risk prediction.
\newblock In {\em Proceedings of the 2018 SIAM International Conference on Data Mining}, pages 261--269. SIAM, 2018.

\bibitem{vaswani2017attention}
Ashish Vaswani, Noam Shazeer, Niki Parmar, Jakob Uszkoreit, Llion Jones, Aidan~N Gomez, {\L}ukasz Kaiser, and Illia Polosukhin.
\newblock Attention is all you need.
\newblock {\em Advances in neural information processing systems}, 30, 2017.

\bibitem{dey2017gate}
Rahul Dey and Fathi~M Salem.
\newblock Gate-variants of gated recurrent unit (gru) neural networks.
\newblock In {\em 2017 IEEE 60th international midwest symposium on circuits and systems (MWSCAS)}, pages 1597--1600. IEEE, 2017.

\bibitem{wang2020c}
L~Wang.
\newblock C-reactive protein levels in the early stage of covid-19.
\newblock {\em Medecine et maladies infectieuses}, 50(4):332--334, 2020.

\bibitem{gao2020dr}
Junyi Gao, Cao Xiao, Lucas~M Glass, and Jimeng Sun.
\newblock Dr. agent: Clinical predictive model via mimicked second opinions.
\newblock {\em Journal of the American Medical Informatics Association}, 27(7):1084--1091, 2020.

\bibitem{huang2020clinical}
Chaolin Huang, Yeming Wang, Xingwang Li, Lili Ren, Jianping Zhao, Yi~Hu, Li~Zhang, Guohui Fan, Jiuyang Xu, Xiaoying Gu, et~al.
\newblock Clinical features of patients infected with 2019 novel coronavirus in wuhan, china.
\newblock {\em The lancet}, 395(10223):497--506, 2020.

\bibitem{zhang2022m3care}
Chaohe Zhang, Xu~Chu, Liantao Ma, Yinghao Zhu, Yasha Wang, Jiangtao Wang, and Junfeng Zhao.
\newblock M3care: Learning with missing modalities in multimodal healthcare data.
\newblock In {\em Proceedings of the 28th ACM SIGKDD Conference on Knowledge Discovery and Data Mining}, KDD '22, page 2418–2428, New York, NY, USA, 2022. Association for Computing Machinery.

\bibitem{weissman2020locally}
Gary~E Weissman, Andrew Crane-Droesch, Corey Chivers, ThaiBinh Luong, Asaf Hanish, Michael~Z Levy, Jason Lubken, Michael Becker, Michael~E Draugelis, George~L Anesi, et~al.
\newblock Locally informed simulation to predict hospital capacity needs during the covid-19 pandemic.
\newblock {\em Annals of internal medicine}, 173(1):21--28, 2020.

\bibitem{ma2021distilling}
Liantao Ma, Xinyu Ma, Junyi Gao, Xianfeng Jiao, Zhihao Yu, Chaohe Zhang, Wenjie Ruan, Yasha Wang, Wen Tang, and Jiangtao Wang.
\newblock Distilling knowledge from publicly available online emr data to emerging epidemic for prognosis.
\newblock In {\em Proceedings of the Web Conference 2021}, pages 3558--3568, 2021.

\bibitem{ma2023aicare}
Liantao Ma, Chaohe Zhang, Junyi Gao, Xianfeng Jiao, Zhihao Yu, Yinghao Zhu, Tianlong Wang, Xinyu Ma, Yasha Wang, Wen Tang, Xinju Zhao, Wenjie Ruan, and Tao Wang.
\newblock Mortality prediction with adaptive feature importance recalibration for peritoneal dialysis patients.
\newblock {\em Patterns}, 4(12), 2023.

\bibitem{mpau}
Mohamed Ragab, Emadeldeen Eldele, Chuan-Sheng Foo, Min Wu, Xiaoli Li, and Zhenghua Chen.
\newblock Source-free domain adaptation with temporal imputation for time series data.
\newblock In {\em 29th SIGKDD Conference on Knowledge Discovery and Data Mining - Research Track}, 2023.

\bibitem{feng2021completing}
Yujie Feng, Jiangtao Wang, Yasha Wang, and Sumi Helal.
\newblock Completing missing prevalence rates for multiple chronic diseases by jointly leveraging both intra- and inter-disease population health data correlations.
\newblock In {\em Proceedings of the Web Conference 2021}, WWW '21, page 183–193, New York, NY, USA, 2021. Association for Computing Machinery.

\bibitem{liu2020neutrophil}
Jingyuan Liu, Yao Liu, Pan Xiang, Lin Pu, Haofeng Xiong, Chuansheng Li, Ming Zhang, Jianbo Tan, Yanli Xu, Rui Song, et~al.
\newblock Neutrophil-to-lymphocyte ratio predicts critical illness patients with 2019 coronavirus disease in the early stage.
\newblock {\em Journal of translational medicine}, 18(1):1--12, 2020.

\bibitem{lopes2021improving}
Ricardo~R Lopes, Hidde Bleijendaal, Lucas~A Ramos, Tom~E Verstraelen, Ahmad~S Amin, Arthur~AM Wilde, Yigal~M Pinto, Bas~AJM de~Mol, and Henk~A Marquering.
\newblock Improving electrocardiogram-based detection of rare genetic heart disease using transfer learning: An application to phospholamban p. arg14del mutation carriers.
\newblock {\em Computers in Biology and Medicine}, 131:104262, 2021.

\bibitem{malhotra2017timenet}
Pankaj Malhotra, Vishnu TV, Lovekesh Vig, Puneet Agarwal, and Gautam Shroff.
\newblock Timenet: Pre-trained deep recurrent neural network for time series classification.
\newblock {\em arXiv preprint arXiv:1706.08838}, 2017.

\bibitem{wardi2021predicting}
Gabriel Wardi, Morgan Carlile, Andre Holder, Supreeth Shashikumar, Stephen~R Hayden, and Shamim Nemati.
\newblock Predicting progression to septic shock in the emergency department using an externally generalizable machine-learning algorithm.
\newblock {\em Annals of emergency medicine}, 77(4):395--406, 2021.

\bibitem{seah2011healing}
Chun-Wei Seah, Ivor Wai-Hung Tsang, and Yew-Soon Ong.
\newblock Healing sample selection bias by source classifier selection.
\newblock In {\em 2011 IEEE 11th International Conference on Data Mining}, pages 577--586. IEEE, 2011.

\bibitem{zhu2023m3fair}
Yinghao Zhu, Jingkun An, Enshen Zhou, Lu~An, Junyi Gao, Hao Li, Haoran Feng, Bo~Hou, Wen Tang, Chengwei Pan, and Liantao Ma.
\newblock M3fair: Mitigating bias in healthcare data through multi-level and multi-sensitive-attribute reweighting method.
\newblock {\em arXiv preprint arXiv:2306.04118}, 2023.

\bibitem{gao2024comprehensive}
Junyi Gao, Yinghao Zhu, Wenqing Wang, Zixiang Wang, Guiying Dong, Wen Tang, Hao Wang, Yasha Wang, Ewen~M Harrison, and Liantao Ma.
\newblock A comprehensive benchmark for covid-19 predictive modeling using electronic health records in intensive care.
\newblock {\em Patterns}, 5(4), 2024.

\bibitem{ma2020concare}
Liantao Ma, Chaohe Zhang, Yasha Wang, Wenjie Ruan, Jiangtao Wang, Wen Tang, Xinyu Ma, Xin Gao, and Junyi Gao.
\newblock Concare: Personalized clinical feature embedding via capturing the healthcare context.
\newblock {\em Proceedings of the AAAI Conference on Artificial Intelligence}, 34(01):833--840, Apr. 2020.

\bibitem{muller2007dynamic}
Meinard M{\"u}ller.
\newblock Dynamic time warping.
\newblock {\em Information retrieval for music and motion}, pages 69--84, 2007.

\bibitem{fawaz2018transfer}
Hassan~Ismail Fawaz, Germain Forestier, and et~al.
\newblock Transfer learning for time series classification.
\newblock In {\em 2018 IEEE international conference on big data (Big Data)}, pages 1367--1376. IEEE, 2018.

\bibitem{reyna2019early}
Matthew~A Reyna, Chris Josef, Salman Seyedi, Russell Jeter, Supreeth~P Shashikumar, M~Brandon Westover, Ashish Sharma, Shamim Nemati, and Gari~D Clifford.
\newblock Early prediction of sepsis from clinical data: the physionet/computing in cardiology challenge 2019.
\newblock In {\em 2019 Computing in Cardiology (CinC)}, pages Page--1. IEEE, 2019.

\bibitem{yan2020interpretable}
Li~Yan, Hai-Tao Zhang, Jorge Goncalves, Yang Xiao, Maolin Wang, Yuqi Guo, Chuan Sun, Xiuchuan Tang, Liang Jing, Mingyang Zhang, et~al.
\newblock An interpretable mortality prediction model for covid-19 patients.
\newblock {\em Nature machine intelligence}, 2(5):283--288, 2020.

\bibitem{hmh}
HM~Hospitales.
\newblock Covid data save lives.
\newblock \url{https://www.hmhospitales.com/prensa/notas-de-prensa/comunicado-covid-data-save-lives}, 2020.
\newblock Accessed: 2025-01-22.

\bibitem{wang2024protocol}
Tianlong Wang, Yinghao Zhu, Zixiang Wang, Wen Tang, Xinju Zhao, Tao Wang, Yasha Wang, Junyi Gao, Liantao Ma, and Ling Wang.
\newblock Protocol to process follow-up electronic medical records of peritoneal dialysis patients to train ai models.
\newblock {\em STAR protocols}, 5(4):103335, 2024.

\bibitem{chai2014root}
Tianfeng Chai and Roland~R Draxler.
\newblock Root mean square error (rmse) or mean absolute error (mae)?--arguments against avoiding rmse in the literature.
\newblock {\em Geoscientific model development}, 7(3):1247--1250, 2014.

\bibitem{fawcett2006introduction}
Tom Fawcett.
\newblock An introduction to roc analysis.
\newblock {\em Pattern recognition letters}, 27(8):861--874, 2006.

\bibitem{zhu2023pyehr}
Yinghao Zhu, Wenqing Wang, Junyi Gao, and Liantao Ma.
\newblock Pyehr: A predictive modeling toolkit for electronic health records.
\newblock \url{https://github.com/yhzhu99/pyehr}, 2023.

\bibitem{baytas2017patient}
Inci~M Baytas, Cao Xiao, Xi~Zhang, Fei Wang, Anil~K Jain, and Jiayu Zhou.
\newblock Patient subtyping via time-aware lstm networks.
\newblock In {\em Proceedings of the 23rd ACM SIGKDD international conference on knowledge discovery and data mining}, pages 65--74, 2017.

\bibitem{ganin2015unsupervised}
Yaroslav Ganin and Victor Lempitsky.
\newblock Unsupervised domain adaptation by backpropagation.
\newblock In {\em International conference on machine learning}, pages 1180--1189. PMLR, 2015.

\bibitem{wilson2020multi}
Garrett Wilson, Janardhan~Rao Doppa, and Diane~J Cook.
\newblock Multi-source deep domain adaptation with weak supervision for time-series sensor data.
\newblock In {\em Proceedings of the 26th ACM SIGKDD international conference on knowledge discovery \& data mining}, pages 1768--1778, 2020.

\end{thebibliography}

\end{document}